# THE 3-TIER INTEGRATED TECHNOLOGY RISK MODEL FOR OCCUPATIONAL SAFETY AND HEALTH

Ramiro Pesina [1] and Ashish Madkaikar [2]

[1] Department of Human Resources Development, University of Texas at Tyler
[2] Department of Technology and Development, Roughrider Consulting

## ABSTRACT

*This paper presents the 3-Tier Integrated Technology Risk Model, a progressive framework for Occupational Safety and Health (OSH) management that integrates Job Safety Analysis (JSA), dynamic risk assessment using leading and lagging indicators, and predictive analytics powered by Large Language Models (LLMs). The model provides organizations with a structured pathway to evolve from static, reactive safety practices to a sophisticated, predictive safety paradigm. Each tier builds upon the previous one, creating a comprehensive, data-driven approach that enables just-in-time training and preemptive risk mitigation.*

## 1. INTRODUCTION

The imperative to ensure a safe and healthy working environment is a cornerstone of modern industrial and corporate governance. For decades, the field of Occupational Safety and Health (OSH) has been guided by principles of hazard identification, risk assessment, and control. Methodologies such as the Job Safety Analysis (JSA) or Job Hazard Analysis (JHA) have become standard practice, providing a structured approach to deconstructing tasks and identifying potential hazards [1]. These methods have undoubtedly contributed to a significant reduction in workplace injuries and fatalities over the years. However, the contemporary workplace is characterized by increasing complexity, technological disruption, and a pace of change that challenges the efficacy of these traditional, often static, risk assessment models [2].

Traditional models excel at identifying static risks; the predictable, unchanging hazards inherent in a specific task or environment. However, they are less adept at managing dynamic risks, which emerge from the complex interplay of human factors, changing environmental conditions, and evolving operational processes [3]. The reliance on lagging indicators, such as incident and injury rates, means that safety interventions are often reactive, implemented only after a negative event has occurred. This reactive posture represents a fundamental limitation in the pursuit of a truly proactive safety culture.

In recent years, the discourse around OSH has shifted towards the adoption of leading indicators and more dynamic forms of risk assessment [4]. Concurrently, the digital transformation sweeping across industries has unlocked unprecedented opportunities to collect and analyze vast quantities of operational data. The convergence of these trends sets the stage for a paradigm shift in safety management. The emergence of advanced analytical technologies, particularly artificial intelligence (AI), machine learning (ML), and most recently, Large Language Models (LLMs), offers the potential to move beyond mere reaction and into the realm of genuine prediction [5, 6]. This paper proposes a conceptual framework, the 3-Tier Integrated Technology Risk Model, that provides a structured pathway for organizations to leverage these technological advancements. The model is designed to be progressive, allowing organizations to build capabilities over time, moving from a foundational, static approach to a sophisticated, predictive one. It integrates established safety methodologies with cutting-edge AI to create a holistic, learning-oriented safety management system.

Tier 1: Static Risk Assessment utilizes the proven JSA methodology as the foundation for identifying and controlling baseline hazards.

Tier 2: Dynamic Risk Analysis introduces a feedback loop, using data from incidents, near-misses, and leading indicators to continuously update and refine the static risk profile.

Tier 3: Predictive Risk Mitigation employs LLMs to synthesize and analyze the comprehensive data from the lower tiers, identifying complex patterns and predicting future risks to enable preemptive interventions.

By framing the evolution of safety management within this three-tiered structure, this paper aims to provide a clear, actionable model for integrating technology into OSH. It specifically addresses the need for more advanced educational and training paradigms, where learning is not a one-time event but a continuous, data-informed process. In doing so, this work contributes to the critical discourse at the intersection of occupational safety, educational technology, and artificial intelligence, offering a vision for a future where workplace incidents are not just managed, but predicted and prevented.

## 2. LITERATURE REVIEW

The development of the 3-Tier Integrated Technology Risk Model is grounded in several distinct but interconnected streams of literature: traditional OSH risk assessment, the shift towards dynamic risk analysis and leading indicators, the role of technology in safety education, and the application of AI and LLMs in predictive analytics.

### 2.1. Traditional Risk Assessment: The Job Safety Analysis (JSA)

The Job Safety Analysis (JSA), often used interchangeably with Job Hazard Analysis (JHA), is a cornerstone of occupational safety programs worldwide. The U.S. Occupational Safety and Health Administration (OSHA) defines it as a process to identify hazards and potential accident causes by breaking a job into its component steps [1]. This systematic examination allows for the implementation of specific control measures for each identified hazard, forming a foundational layer of an organization’s safety management system. The strength of the JSA lies in its structured, methodical nature, which provides a clear and documented assessment of known risks associated with routine tasks. Research has consistently demonstrated the utility of JSA in various industries, from construction and manufacturing to healthcare, as a primary tool for hazard identification and static risk assessment [7, 8].

However, the very structure that makes JSA effective also contributes to its primary limitation: it is inherently static. A JSA is typically performed at a single point in time and may not be regularly updated to reflect changes in procedures, equipment, or personnel. This static nature can lead to a“false sense of security,” where organizations believe risks are controlled, while new, unassessed hazards emerge [9]. This gap highlights the need for a more dynamic approach to complement the foundational static analysis provided by the JSA.

## 2.2. The Shift to Dynamic Risk and Leading Indicators

Recognizing the limitations of static assessments, the OSH field has increasingly emphasized the importance of dynamic risk assessment. Unlike static risk, which is predictable and stable, dynamic risk is fluid and context-dependent, influenced by real-time operational conditions [3]. The management of dynamic risk requires continuous monitoring and adaptation, moving beyond a compliance-based mindset to one of constant vigilance.

This shift is closely tied to the advocacy for leading indicators in safety performance measurement. While lagging indicators (e.g., incident rates) measure failures that have already occurred, leading indicators are proactive measures that provide insight into the effectiveness of safety and health activities before an incident happens [4]. Examples of leading indicators include the frequency of safety observations, the percentage of employees completing safety training, and the time taken to resolve identified hazards. OSHA has published extensive guidance on using leading indicators to improve safety outcomes, arguing that they are essential for creating a proactive safety culture [4].

The integration of leading and lagging indicators creates a feedback loop that can inform and enhance traditional risk assessments. By analyzing trends in both types of indicators, organizations can identify areas where static controls may be failing or where new hazards are emerging. This iterative process of analysis and refinement forms the conceptual basis for Tier 2 of the proposed model, transforming the static JSA into a living document that evolves with the organization.

## 2.3. Technology in Safety Education and Human Resource Development (HRD)

The effective management of OSH is fundamentally an educational endeavor, deeply intertwined with Human Resource Development (HRD). Training is a critical component of any safety program, yet traditional, one-size-fits-all training models often fail to engage employees or address the specific risks they face. The integration of technology into safety education offers a pathway to more personalized, effective, and continuous learning experiences.

Immersive technologies like virtual reality (VR) and augmented reality (AR) have shown significant promise in providing realistic, hands-on safety training without exposing employees to actual hazards [10]. E-learning platforms and Learning Management Systems (LMS) allow for the flexible delivery and tracking of safety training, ensuring that employees can access information when and where they need it [11]. From an HRD perspective, integrating OSH into broader human resource management practices, such as performance management and leadership development, is crucial for fostering a strong safety culture [12, 13]. Technology can facilitate this integration by providing the data and analytics needed to identify training needs and measure the impact of safety initiatives on organizational performance.

## 2.4. Predictive Analytics: The Role of AI and Large Language Models (LLMs)

The most recent and transformative development in the technological landscape is the rise of AI and LLMs. These technologies have the capacity to analyze vast and complex datasets, uncovering patterns and making predictions that are beyond human capability. In the context of OSH, predictive analytics can be used to forecast the likelihood of future incidents by analyzing historical data on incidents, near-misses, safety observations, and operational conditions [5].

Early applications of machine learning in safety have demonstrated success in predicting injury risks with a high degree of accuracy [14]. The advent of LLMs opens up new frontiers. LLMs can process and understand unstructured data, such as text from incident reports, safety meeting minutes, and employee feedback, in addition to structured numerical data. This allows for a more holistic analysis of the safety environment. Recent studies have explored the use of LLMs for tasks such as construction safety risk assessment and analyzing safety reports, highlighting their potential to revolutionize safety practices [6, 15]. By synthesizing all available information, LLMs can identify subtle, precursor patterns that signal an increased risk of an incident, forming the core of the predictive capability in Tier 3 of the proposed model.

This review of the literature establishes the foundation for the 3-Tier Integrated Technology Risk Model. It brings together the established principles of JSA, the proactive philosophy of dynamic risk assessment and leading indicators, the educational focus of HRD, and the transformative power of AI and LLMs. The proposed model integrates these elements into a cohesive framework designed to guide organizations on a journey of continuous safety improvement.

## 3. THE 3-TIER INTEGRATED TECHNOLOGY RISK MODEL

The 3-Tier Integrated Technology Risk Model is a conceptual framework designed to guide organizations in systematically enhancing their OSH management systems. It provides a structured, progressive pathway from foundational, static risk assessment to a sophisticated, predictive safety paradigm. Each tier builds upon the last, creating an increasingly comprehensive and data-driven approach to safety. The model is not intended as a rigid prescription but as a flexible architecture that can be adapted to the specific context and maturity level of any organization.

### 3.1. Tier 1: Static Risk Assessment – The Foundation

Objective: To identify and control known, predictable hazards associated with specific jobs or tasks.

Core Methodology: Job Safety Analysis (JSA) / Job Hazard Analysis (JHA).

Tier 1 represents the essential starting point for any systematic safety program. It is focused on establishing a baseline understanding of the inherent risks within the organization. The primary tool for this tier is the JSA. The process involves:

Task Decomposition: Breaking down a job into a sequence of discrete steps.

Hazard Identification: For each step, identifying potential hazards to workers and the environment.

Control Development: Devising preventive measures to eliminate or mitigate the identified hazards. These controls can follow the hierarchy of controls, from elimination and substitution to engineering controls, administrative controls, and finally, personal protective equipment (PPE).

Role of Technology: At this foundational tier, technology plays a role in digitizing and managing the JSA process. Digital JSA platforms can streamline the creation, storage, and retrieval of analyses. They can ensure consistency in format and provide easy access for employees and supervisors. This digital repository of static risks is the foundational dataset upon which the subsequent tiers are built.

Output: A comprehensive library of JSAs that documents the known, static hazards across the organization and the established control measures. This tier ensures compliance with fundamental OSH regulations and establishes a baseline safety culture.

### 3.2. Tier 2: Dynamic Risk Analysis – The Feedback Loop

Objective: To move beyond a static view of risk by incorporating real-time operational data and creating a continuous improvement cycle.

Core Methodology: Integration of Lagging and Leading Indicators with JSA.

Tier 2 addresses the primary limitation of Tier 1: its static nature. This tier introduces a dynamic feedback loop that allows the organization to learn from its daily operations and adapt its risk assessments accordingly. The core of Tier 2 is the systematic collection and analysis of both lagging and leading indicators.

Lagging Indicators: These are reactive measures of safety failures. They include data from:
Incident reports (injuries, illnesses, fatalities)
Near-miss reports
Property damage reports
First-aid logs

Leading Indicators: These are proactive measures of safety performance. They include data from:

- Safety observations and inspections
- Safety audit results
- Completion rates for safety training
- Employee perception surveys
- Hazard reporting and resolution tracking

The Feedback Mechanism: The key to Tier 2 is the process of analyzing these indicators to identify trends and patterns. For example, a rising trend in near-misses related to a specific task might indicate that the controls defined in the Tier 1 JSA are ineffective or not being followed. Conversely, a high rate of completed safety observations (a leading indicator) might correlate with a decrease in incidents (a lagging indicator).

Role of Technology: Technology is crucial for the effective functioning of Tier 2. Safety management software platforms can be used to collect, aggregate, and visualize data from various sources. Dashboards can provide real-time insights into safety performance, allowing managers to track trends in leading and lagging indicators. This technology enables the iterative refinement of the JSAs. When a trend suggests a new or uncontrolled risk, the relevant JSA is reviewed and updated with new control measures. This transforms the static JSA of Tier 1 into a dynamic, evolving document.

Output: A continuously updated risk profile for the organization. Tier 2 fosters a learning culture where safety is not a static set of rules but a dynamic process of monitoring, analyzing, and adapting. It moves the organization from a purely reactive to a more proactive safety posture.

### 3.3. Tier 3: Predictive Risk Mitigation – The Foresight Engine

Objective: To leverage the full spectrum of available data to predict future safety risks and enable preemptive interventions.

Core Methodology: Application of Artificial Intelligence (AI) and Large Language Models (LLMs).

Tier 3 represents the most advanced stage of the model, where the organization moves from being proactive to being predictive. This tier harnesses the power of AI to analyze the vast and complex datasets generated by Tiers 1 and 2. The goal is to identify subtle, non-obvious patterns that are precursors to safety incidents.

The Predictive Engine: The core of Tier 3 is an AI model, likely incorporating LLMs, that is trained on all available safety and operational data. This includes:

- The entire library of digital JSAs (Tier 1).
- All structured data from leading and lagging indicators (Tier 2).
- Unstructured text data from incident reports, witness statements, safety meeting minutes, and open-ended comments from employee surveys.
- Potentially, operational data from other business systems, such as production schedules, maintenance logs, and environmental sensor data.

By processing this rich, multimodal dataset, the LLM can learn the complex relationships between various factors and the occurrence of safety incidents. For example, the model might discover that a combination of a specific task, a recent equipment maintenance event, an inexperienced operator, and a high-production-pressure day creates a significantly elevated risk of a particular type of incident, a pattern that would be nearly impossible for a human analyst to detect.

Role of Technology: This tier is entirely dependent on advanced AI technology. It requires a robust data infrastructure to collect and process the data, as well as the expertise to build, train, and validate the predictive models. The LLM acts as a“synthesis engine,” capable of understanding the nuances of human language in safety reports and combining that understanding with quantitative data to generate a holistic risk forecast [15].

Output: The primary output of Tier 3 is a predictive risk forecast. This is not a deterministic prediction of exactly when and where an accident will occur, but rather a probabilistic assessment of heightened risk. This forecast can be presented through a dashboard that flags high-risk scenarios, tasks, or locations, allowing safety professionals and operational managers to take preemptive action. These actions could include deploying additional safety supervision, temporarily halting a high-risk operation for a safety review, or, most powerfully, triggering targeted, just-in-time training interventions for the specific employees involved.

# 4. Conceptual Model Architecture And Information Flow

The synergy between the three tiers is what gives the model its power. The architecture is designed for a continuous and evolving flow of information, where each tier informs and enhances the others. A visual representation of this architecture is presented in Figure 1, followed by a detailed explanation of the information pathways.

## 4.1. Information Flow from Tier 1 to Tier 2

The flow of information begins with the foundational data established in Tier 1. The digital library of JSAs provides the initial, static risk profile for the organization. This repository serves as the baseline against which the dynamic data of Tier 2 is compared. When a safety incident or near-miss occurs (a lagging indicator), the report is cross-referenced with the relevant JSA. The analysis seeks to answer critical questions: Was a JSA in place for this task? Were the identified hazards accurate? Were the prescribed controls implemented and effective? This direct comparison allows for a data-driven refinement of the JSA. If the analysis reveals a deficiency in the original JSA, it is updated. This process ensures that the organization learns from its failures and that the static risk assessments do not become obsolete.

## 4.2. Information Flow from Tier 2 to Tier 3

Tier 2 generates a rich stream of both structured and unstructured data. The structured data includes quantitative metrics from leading and lagging indicators (e.g., number of safety observations, incident frequency rates). The unstructured data includes the narrative text from incident reports, witness statements, and safety meeting notes. This entire dataset, representing the dynamic safety performance of the organization, is fed into the Tier 3 predictive engine. The AI and LLMs in Tier 3 require this comprehensive and continuous flow of information to learn the complex patterns of risk. The more data the model receives, the more accurate its predictions become. Tier 2 essentially acts as the data-gathering and initial processing layer for the advanced analytics of Tier 3.

## 4.3. The Predictive Feedback Loop (Tier 3 to Operations)

The output of Tier 3, the predictive risk forecast, completes the cycle. This forecast is not a passive report; it is an actionable intelligence product. When the model flags a high-risk scenario, it triggers a response. This response is directed back at the operational level. For example, if the model predicts an elevated risk of a fall from height on a specific construction site due to a combination of factors, a notification can be sent to the site supervisor. This allows for immediate, targeted interventions, such as a pre-task safety briefing focused on fall protection or an additional inspection of scaffolding. This predictive feedback loop is the ultimate goal of the model: to use foresight to prevent incidents before they occur.

# 5. Implications for Educational Technology and Human Resource Development

The 3-Tier Integrated Technology Risk Model is not merely a framework for safety management; it is a blueprint for a new paradigm of safety education and training. By integrating technology at each level, the model transforms safety learning from a static, compliance-driven requirement into a dynamic, data-informed, and continuous process. This aligns perfectly with the goals of both educational technology and modern Human Resource Development (HRD), which seek to create more effective, engaging, and impactful learning experiences.

## 5.1. From Static Training to Dynamic Learning

Traditional safety training is often a one-time event, where employees attend a class or watch a video. This approach is disconnected from the realities of the workplace and suffers from low knowledge retention. The 3-Tier Model facilitates a shift towards a dynamic learning ecosystem: Tier 1 provides the foundational curriculum. The digital JSAsserve as core educational content, clearly outlining the known hazards and control measures for specific tasks. This ensures that all employees have access to a baseline of essential safety knowledge.

Tier 2 introduces a continuous learning element. The analysis of incident and near-miss data provides real-world case studies that can be used for training. When a JSA is updated based on new data, this becomes a learning opportunity. Micro-learning modules can be deployed to inform employees of the changes and the reasons behind them. This process of continuous refinement reinforces the idea that safety is an evolving field and that learning is an ongoing responsibility.

## 5.2. The Advent of Predictive, Just-in-Time Training

The most profound educational implication of the model lies in Tier 3. The predictive capabilities of the AI engine enable a move towards just-in-time, personalized safety training. Instead of training everyone on every possible hazard, organizations can deliver targeted training to the right people at the right time.

When the Tier 3 model predicts a heightened risk for a specific employee or team, it can automatically trigger a training intervention. For example:

If the LLM identifies that an employee who has not recently been trained on lockout/tagout procedures is scheduled to perform maintenance on a piece of equipment that has a history of unexpected startups, it could automatically assign a brief refresher e-learning module to that employee's training queue. The system could even prevent the employee from starting the task until the training is complete.

This represents a true integration of educational technology with operational systems. It is a shift from a "just-in-case" training model to a "just-in-time" learning model, where education is delivered at the moment of need. This approach is not only more efficient but also far more effective, as the learning is directly relevant to the immediate context of the employee's work.

## 5.3. Fostering a Data-Driven Learning Culture

From an HRD perspective, the 3-Tier Model provides the framework for building a genuine learning culture focused on safety. By making safety data transparent and accessible (through the dashboards in Tier 2 and the forecasts in Tier 3), the model empowers employees to take ownership of their safety and the safety of their colleagues. It moves the conversation from one of blame (after an incident) to one of foresight and continuous improvement.

This data-driven approach also allows HRD professionals to demonstrate the value of safety training in a more tangible way. By correlating training interventions with changes in leading and lagging indicators, they can measure the return on investment (ROI) of their programs and make a stronger business case for continued investment in safety education [16].

# 6. IMPLEMENTATION FRAMEWORK AND ORGANIZATIONAL READINESS

The successful implementation of the 3-Tier Integrated Technology Risk Model requires careful planning, organizational readiness, and a phased approach. This section provides a practical framework for organizations seeking to adopt the model, addressing the key considerations at each stage of implementation.

## 6.1. Assessing Organizational Maturity

Before embarking on the implementation journey, organizations must conduct an honest assessment of their current safety management maturity. This assessment should evaluate several dimensions:

Data Infrastructure: Does the organization have systems in place to collect, store, and manage safety data? Are incident reports digitized? Is there a centralized repository for JSAs? Organizations with paper-based systems will need to invest in digitization before progressing beyond Tier 1.

Safety Culture: Is there a culture of reporting and learning from incidents and near-misses? Do employees feel comfortable raising safety concerns without fear of retribution? A punitive culture will undermine the data collection efforts essential for Tiers 2 and 3.

Technological Capability: Does the organization have the in-house expertise to implement and manage advanced data analytics and AI systems? If not, are there resources available to partner with external vendors or consultants?

Leadership Commitment: Is there visible and sustained commitment from senior leadership to invest in safety technology and to support the cultural changes required? Without this commitment, implementation efforts are likely to stall.

This maturity assessment will help organizations determine their starting point and set realistic timelines for progression through the tiers.

## 6.2. Phased Implementation Strategy

The 3-Tier Model is designed to be implemented in phases, allowing organizations to build capabilities incrementally.

Phase 1: Establishing the Foundation (Tier 1)

The first phase focuses on building a comprehensive, digital library of JSAs. This involves:

- Standardization: Developing a standardized JSA template and process across the organization.
- Digitization: Implementing a digital platform for creating, storing, and accessing JSAs.
- Training: Training supervisors and employees on how to conduct and use JSAs effectively.
- Completion: Conducting JSAs for all critical tasks and high-risk activities.

This phase typically takes 6-12 months, depending on the size and complexity of the organization. The key deliverable is a complete, accessible digital repository of static risk assessments.

Phase 2: Building the Feedback Loop (Tier 2)

Once a solid foundation is in place, the organization can move to Tier 2. This phase involves:

- Data Integration: Integrating incident reporting systems, safety observation platforms, and training management systems into a unified safety data warehouse.
- Indicator Development: Defining and tracking a comprehensive set of leading and lagging indicators.
- Analytics Implementation: Deploying dashboards and analytics tools to visualize trends and identify patterns.
- Process Establishment: Creating a formal process for reviewing indicator trends and updating JSAs based on the findings.

This phase typically takes 12-18 months. The key deliverable is a functioning feedback loop that continuously refines the organization's understanding of its risk profile.

Phase 3: Achieving Predictive Capability (Tier 3)

The final phase is the most technologically demanding. It involves:

- Data Preparation: Cleaning and structuring the accumulated data to make it suitable for AI training.
- Model Development: Working with data scientists to develop and train predictive models, potentially using LLMs.
- Validation: Rigorously testing the models to ensure they are accurate and do not introduce bias.
- Integration: Integrating the predictive outputs into operational workflows and decision-making processes.
- Governance: Establishing clear governance structures and ethical guidelines for the use of predictive analytics.

This phase can take 18-24 months or longer. The key deliverable is a validated, operational predictive risk system that provides actionable foresight.

## 6.3. Key Success Factors

Several factors are critical to the success of the implementation:

Cross-Functional Collaboration: The implementation requires collaboration between safety professionals, IT departments, HR, operations, and senior leadership. Siloed approaches will fail.
Change Management: The introduction of new technologies and processes requires effective change management. Employees need to understand the "why" behind the changes and be given the support to adapt.

Continuous Improvement: The model itself should be subject to continuous improvement. Organizations should regularly review the effectiveness of each tier and make adjustments as needed.

Pilot Programs: Before full-scale deployment, organizations should consider pilot programs in specific departments or locations. This allows for learning and refinement before broader rollout.

## 7. CONCEPTUAL APPLICATION SCENARIOS

To illustrate the practical application of the 3-Tier Model, this section presents conceptual scenarios across different industries, demonstrating how the model could be adapted to various organizational contexts. These scenarios are theoretical constructs designed to clarify the model's potential implementation, not descriptions of actual past implementations.

### 7.1. Conceptual Scenario 1: Manufacturing Plant – Reducing Machine-Related Injuries

Context: Consider a large automotive manufacturing plant that experiences injuries related to machine operation, particularly during equipment changeovers.

Tier 1 Application: In this conceptual application, the plant would conduct detailed JSAs for all machine operation and changeover procedures. These JSAs identify hazards such as pinch points, moving parts, and lockout/tagout requirements. Digital JSAsare made accessible to operators via tablets at each workstation.

Tier 2 Application: The plant would implement a near-miss reporting app and track leading indicators such as the number of lockout/tagout audits completed and the percentage of operators who have completed refresher training. Analysis could reveal trends such as near-misses related to machine changeovers spiking during periods of high production pressure. The JSAsfor changeover procedures would be updated to include additional checks and a mandatory pre-task safety briefing when production targets are elevated.

Tier 3 Application: An LLM-based predictive model would be trained on historical incident data, production schedules, maintenance logs, and operator experience levels. The model could identify patterns such as: when a specific machine has undergone maintenance in the past 48 hours, and an operator with less than two years of experience is assigned to it during a high-production shift, the risk of a machine-related injury increases significantly. The system would automatically flag this scenario and trigger a notification to the shift supervisor, who could then assign a more experienced operator to mentor the less experienced one during the first changeover.

Anticipated Outcome: Under this conceptual framework, the plant would be positioned to achieve measurable reductions in machine-related injuries, particularly during high-risk scenarios identified by the predictive model.

### 7.2. Conceptual Scenario 2: Construction Site – Preventing Falls from Height

Context: Consider a construction company working on high-rise building projects that seeks to reduce the incidence of falls from height, a leading cause of fatalities in the industry.

Tier 1 Application: In this conceptual framework, the company would develop comprehensive JSAs for all work-at-height activities, including scaffolding erection, steel erection, and roofing. These JSAs detail the required fall protection equipment and procedures.

Tier 2 Application: The company would use a mobile app for daily safety observations and track leading indicators such as the percentage of workers observed using fall protection correctly and the number of fall protection equipment inspections completed. Analysis of incident and near-miss data could reveal that falls are more likely to occur on Mondays and after periods of inclement weather. The company would implement a policy of enhanced safety briefings and additional supervisory presence on Mondays and after weather delays.

Tier 3 Application: A predictive model would analyze data from multiple sources, including weather forecasts, project schedules, crew composition, and safety observation data. The model could predict that on a specific upcoming day, due to a combination of factors (recent rain making surfaces slippery, a new subcontractor crew starting work, and a tight deadline), the risk of a fall from height is elevated. The project manager would receive an alert and could deploy additional safety personnel to the site and conduct a mandatory toolbox talk on fall protection before work begins.

Anticipated Outcome: Under this conceptual application, the company would be positioned to achieve significant reductions in fall-related incidents, with predictive alerts serving as a valued tool for proactive risk management.

### 7.3. Conceptual Scenario 3: Healthcare Facility – Reducing Needlestick Injuries

Context: Consider a large hospital that seeks to reduce the rate of needlestick injuries among nursing staff, which pose a risk of bloodborne pathogen transmission.

Tier 1 Application: In this conceptual application, the hospital would conduct JSAs for all procedures involving sharps, such as IV insertion, blood draws, and medication administration. These JSAs emphasize the use of safety-engineered sharps devices and proper disposal techniques.

Tier 2 Application: The hospital would implement a confidential needlestick injury reporting system and track leading indicators such as the percentage of safety-engineered devices in use and the completion rates for annual bloodborne pathogen training. Analysis could reveal that needlestick injuries are more frequent in the emergency department during night shifts. The hospital would increase staffing levels during these periods and provide additional training focused on the specific challenges of the ED environment.

Tier 3 Application: A predictive model would be developed that analyzes data including shift schedules, patient acuity levels, staffing ratios, and individual nurse experience levels. The model could identify that when a nurse is working a double shift in a high-acuity unit, the risk of a needlestick injury increases. The system would alert the charge nurse, who could then ensure that the nurse takes adequate breaks and is not assigned to the most complex procedures during the latter part of the shift.

Anticipated Outcome: Under this conceptual framework, the hospital would be positioned to achieve sustained reductions in needlestick injuries, protecting the health of its staff and reducing costs associated with post-exposure prophylaxis and testing.

These conceptual scenarios illustrate the potential versatility and practical value of the 3-Tier Model across diverse industries. In each case, the model provides a structured conceptual approach to moving from basic hazard identification to sophisticated, predictive risk management. While these scenarios are theoretical, they are grounded in the realities of each industry and demonstrate the model's adaptability to different operational contexts.

# 8. DISCUSSION

The 3-Tier Integrated Technology Risk Model offers a forward-looking vision for the future of OSH, but its implementation is not without challenges. This section discusses the potential benefits and limitations of the model, explores the broader implications for the field, and suggests directions for future research.

## 8.1. Theoretical Contributions

Before examining the practical implications, it is important to acknowledge the theoretical contributions of the 3-Tier Model to the field of OSH and educational technology. The model synthesizes several theoretical frameworks:

Systems Theory: The model views the organization as a complex system where safety is an emergent property of the interactions between people, processes, technology, and the environment. Each tier of the model represents a different level of system understanding, from the component level (Tier 1) to the system dynamics level (Tier 2) to the predictive system behavior level (Tier 3).

Learning Organization Theory: Drawing on the work of Senge and others, the model embodies the principles of a learning organization. It creates structures for continuous learning (Tier 2) and uses that learning to adapt and improve (Tier 3). The feedback loops are mechanisms for organizational learning at scale.

Technology Acceptance and Integration: From an educational technology perspective, the model provides a structured approach to technology integration that addresses both the pedagogical and the organizational dimensions. It recognizes that technology is not a panacea but a tool that must be thoughtfully integrated into existing practices and cultures.

These theoretical underpinnings provide a solid foundation for the model and situate it within the broader academic discourse.

## 8.2. Significance and Potential Benefits

The primary significance of the model is its potential to fundamentally shift the paradigm of safety management from a reactive to a predictive stance. By systematically building upon foundational safety practices with layers of data analysis and artificial intelligence, the model provides a clear pathway for organizations to mature their safety programs. The potential benefits are substantial:

Reduced Incidents: By predicting and preempting hazardous situations, the model has the direct potential to reduce the frequency and severity of workplace incidents.

Enhanced Learning: The model transforms safety training into a continuous, personalized, and highly relevant learning experience, which can lead to better knowledge retention and safer work practices.

Improved Efficiency: Just-in-time training and targeted interventions are more efficient than broad, one-size-fits-all approaches, saving time and resources.

Data-Driven Decision Making: The model provides safety professionals and organizational leaders with the data and insights they need to make more informed decisions about resource allocation and safety strategy.

Strengthened Safety Culture: By fostering transparency, learning, and proactive engagement, the model can help to build a deeply embedded culture of safety.

## 8.3. Limitations and Challenges

Despite its potential, the implementation of the 3-Tier Model faces several significant challenges: Data Requirements: The effectiveness of the model, particularly Tier 3, is entirely dependent on the availability of large volumes of high-quality data. Organizations with poor data collection practices will struggle to implement the more advanced tiers.

Technological Barriers: The AI and LLM technologies at the heart of Tier 3 are complex and require specialized expertise to develop, implement, and maintain. The cost of this technology and the talent required to manage it may be prohibitive for smaller organizations.

Ethical Considerations: The use of AI to monitor and predict employee behavior raises important ethical questions. Issues of privacy, surveillance, and algorithmic bias must be carefully addressed. For example, if a predictive model disproportionately flags a certain demographic group, it could lead to discriminatory practices. Transparency in how the models work and clear governance structures are essential.

Over-Reliance on Technology: There is a risk that an over-reliance on the predictive model could lead to a diminishment of human oversight and professional judgment. The model should be seen as a tool to support, not replace, the expertise of safety professionals.

## 8.4. The Role of Stakeholders in Model Implementation

The successful deployment of the 3-Tier Model requires the active engagement of multiple stakeholders, each playing a distinct but interconnected role.

Safety Professionals: Safety professionals are the primary stewards of the model. They are responsible for conducting JSAs, analyzing safety data, interpreting predictive outputs, and recommending interventions. The model enhances their role by providing them with more powerful tools and data-driven insights, but it also requires them to develop new competencies in data analytics and technology management.

Frontline Employees: Employees are both the beneficiaries and the contributors to the model. They benefit from improved safety through better hazard identification and proactive interventions. They contribute by reporting incidents and near-misses, participating in safety observations, and providing feedback on the effectiveness of controls. For the model to succeed, employees must trust that the data will be used for learning and improvement, not for punishment.

Supervisors and Managers: Supervisors and managers are the critical link between the data and action. They receive alerts from the predictive system and are responsible for implementing the recommended interventions. They also play a key role in fostering a culture of safety and encouraging employee participation in data collection.

Human Resource Development (HRD) Professionals: HRD professionals are responsible for designing and delivering the training interventions that are triggered by the model. They must work closely with safety professionals to ensure that training is relevant, timely, and effective. The model provides HRD with valuable data on training needs and the opportunity to demonstrate the impact of their programs.

Information Technology (IT) Professionals: IT professionals provide the technical infrastructure that underpins the model. They are responsible for implementing the data systems, ensuring data security and privacy, and supporting the deployment of AI models. Their expertise is essential, particularly for Tier 3.

Senior Leadership: Senior leadership provides the vision, resources, and commitment necessary for the model to succeed. They must champion the initiative, allocate budget, and hold the organization accountable for continuous improvement in safety performance.

The alignment and collaboration of these stakeholders is not automatic; it requires deliberate effort, clear communication, and a shared understanding of the model's goals and benefits.

## 8.5. Broader Implications for the Field of OSH

Beyond the specific benefits to individual organizations, the widespread adoption of the 3-Tier Model could have profound implications for the field of OSH as a whole.

Shift in Professional Competencies: The model signals a shift in the competencies required of safety professionals. Traditional skills in hazard identification and regulatory compliance remain important, but they must be augmented with skills in data analysis, technology management, and change leadership. Educational programs for safety professionals will need to adapt to prepare graduates for this new reality.

Evolution of Regulatory Frameworks: Current OSH regulations are largely prescriptive, specifying the controls that must be in place for specific hazards. As predictive technologies mature, there may be an opportunity for a shift towards more performance-based regulations, where organizations are given greater flexibility in how they manage risk, provided they can demonstrate, through data, that they are achieving superior safety outcomes.

Industry Collaboration and Data Sharing: The effectiveness of predictive models improves with the volume and diversity of data. There is potential for industry-wide collaboration, where organizations share anonymized safety data to build more robust predictive models. This would require overcoming competitive concerns and establishing trusted data-sharing platforms, but the potential benefits to the entire industry could be substantial.

Ethical and Legal Frameworks: The use of AI in safety management raises new ethical and legal questions that the field must grapple with. Issues of algorithmic transparency, data privacy, worker surveillance, and liability in the event of an AI-predicted incident that is not prevented all require careful consideration and the development of appropriate frameworks.

## 8.6. Directions for Future Research

The 3-Tier Model is a conceptual framework that requires empirical validation. Future research should focus on several key areas:

Pilot Implementation: A pilot study implementing the model in a real-world organizational setting would be invaluable. Such a study could measure the model's impact on leading and lagging safety indicators and provide insights into the practical challenges of implementation.

Model Validation: Research is needed to develop and validate the predictive algorithms used in Tier 3. This would involve testing different AI and LLM architectures to determine which are most effective at predicting safety risks in various industrial contexts.

Ethical Frameworks: There is a need for research into the development of ethical guidelines and governance frameworks for the use of predictive analytics in OSH. This research should involve a multi-stakeholder dialogue, including workers, unions, employers, and technology developers.

Integration with MES: While this paper has separated the detailed discussion of MES integration, future research should explicitly explore the technical and pedagogical challenges of linking the predictive outputs of Tier 3 to automated training triggers within Manufacturing Execution Systems. This would represent the ultimate realization of the just-in-time learning concept.

## 9. CONCLUSION

The traditional approaches to occupational safety and health, while valuable, are no longer sufficient to address the complexities of the modern workplace. The future of safety lies in the intelligent integration of technology to move beyond reaction and towards prediction. The 3-Tier Integrated Technology Risk Model presented in this paper offers a structured and progressive framework for achieving this vision. By building upon the foundational principles of Job Safety Analysis with a dynamic, data-driven feedback loop and the predictive power of artificial intelligence, the model provides a pathway for organizations to create a truly proactive and learning-oriented safety culture.

This paper has presented a comprehensive exploration of the model, from its theoretical foundations to its practical implementation. The literature review established that while traditional JSA methods provide a solid foundation for static risk assessment, they must be complemented by dynamic approaches that incorporate leading and lagging indicators. The integration of advanced AI technologies, particularly Large Language Models, represents the next frontier, enabling organizations to move from reactive and proactive stances to a genuinely predictive paradigm.

The conceptual architecture of the 3-Tier Model provides a clear roadmap for this evolution. Tier 1 establishes the baseline through comprehensive JSA documentation. Tier 2 introduces the critical feedback loop, transforming static assessments into living documents that evolve with organizational learning. Tier 3 harnesses the power of AI to synthesize vast datasets and generate actionable predictive insights. The information flow between these tiers creates a synergistic system where each level enhances the others, resulting in a safety management capability that is greater than the sum of its parts.

The implications for educational technology are profound and multifaceted. The model enables a fundamental shift from static, compliance-based training to a dynamic ecosystem of personalized, just-in-time learning. It transforms safety education from a periodic requirement into a continuous, data-informed process that is woven into the fabric of daily operations. From an HRD perspective, this represents a significant advancement in how organizations develop and deploy their human capital in the service of safety. The model provides the data infrastructure and analytical capabilities needed to demonstrate the ROI of safety training, making a compelling business case for continued investment in human development.

The implementation framework outlined in this paper provides practical guidance for organizations at various stages of maturity. The phased approach recognizes that not all organizations are ready to immediately deploy advanced AI systems, but that they can build towards that capability over time. The application scenarios across manufacturing, construction, and healthcare demonstrate the versatility of the model and its potential to deliver tangible safety improvements across diverse industries.

However, the path forward is not without obstacles. The challenges of data quality, technological investment, ethical governance, and cultural change are significant and must be addressed with care and intentionality. Organizations must resist the temptation to view technology as a silver bullet and instead recognize it as a powerful tool that must be thoughtfully integrated into existing practices and cultures. The role of human judgment, professional expertise, and ethical oversight remains paramount, even as AI capabilities advance.

Looking to the future, the continued evolution of AI and LLM technologies promises even greater capabilities for predictive safety management. As these technologies become more accessible and affordable, the barriers to implementation will lower, potentially democratizing access to advanced safety analytics. The research agenda outlined in this paper provides a roadmap for the empirical validation and refinement of the model, ensuring that its development is grounded in evidence and responsive to the real-world needs of organizations and workers.

Ultimately, the 3-Tier Integrated Technology Risk Model is more than a framework for safety management; it is a vision for a future where work is fundamentally safer because organizations have the foresight to prevent incidents before they occur. It is a vision where learning is continuous, where data drives decisions, and where technology serves humanity by protecting the health and well-being of workers. The journey from static to predictive risk is a journey of organizational learning and technological integration, and it is a journey that is now, more than ever, within our reach. As we stand at the threshold of this new era in occupational safety and health, the question is not whether we can achieve this vision, but whether we have the collective will and commitment to make it a reality.

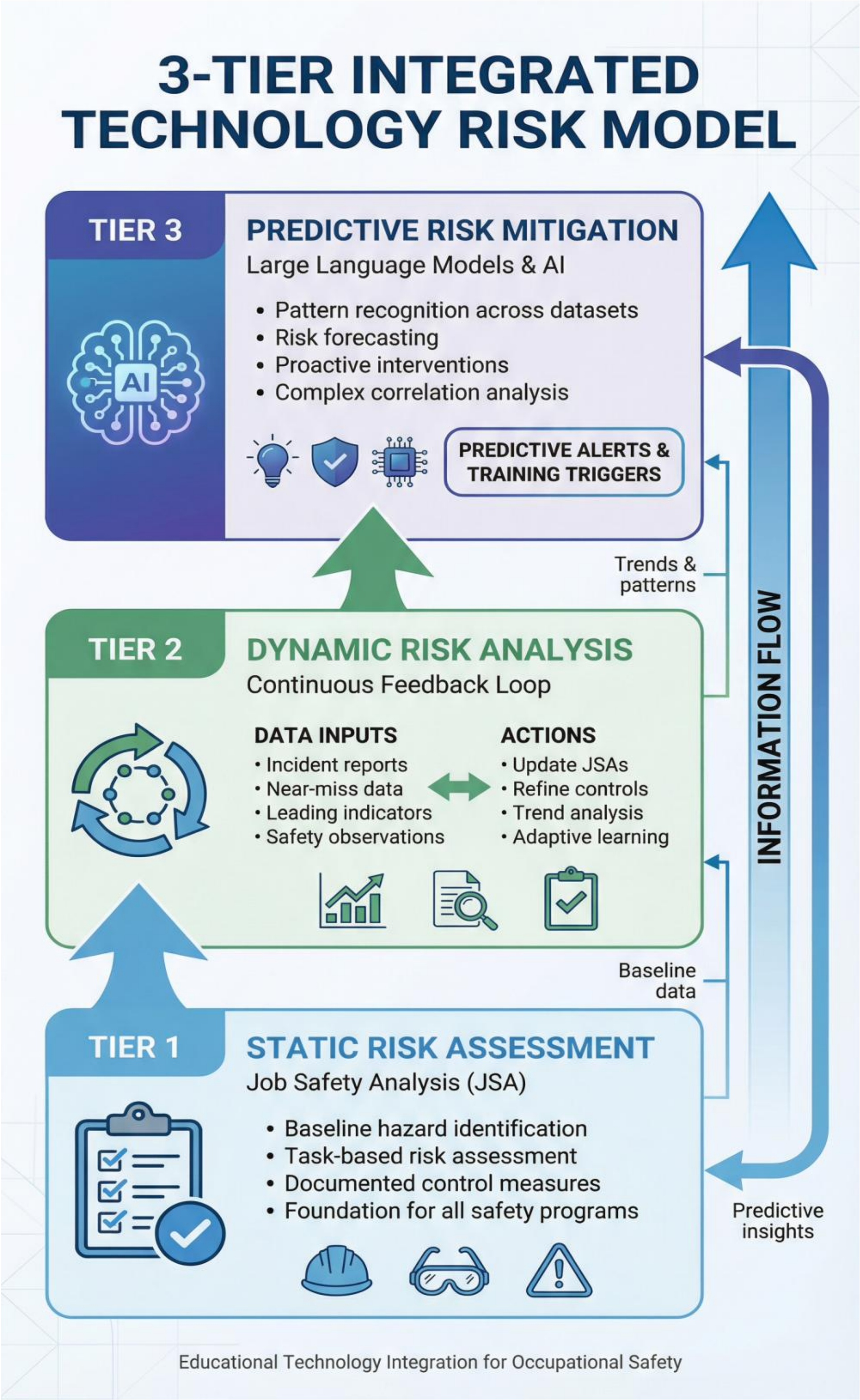

Figure 1. The 3-Tier Integrated Technology Risk Model

## AUTHORS

**Ramiro Pesina, PhD.** is a seasoned leader with over 25 years of experience in safety, sustainability, and environmental management. He currently serves as a Global Director for Environmental, Health, Safety, and Sustainability at Emerson Electric, a role he has held since 2016. In this position, Dr. Pesina is responsible for developing and implementing strategies that foster an award-winning team and an innovative culture.

In addition to his corporate role, Dr. Pesina is pending graduation at the University of Texas at Tyler. His academic research focuses on critical areas such as organizational development, interventions, occupational safety, and workforce development frameworks. He holds an MBA, an MS, and is a Certified Safety Professional (CSP). Dr. Pesina is a published author with research featured in academic journals and is a recognized voice in the fields of occupational safety, organizational mentorship and change management.

**Ashish Madkaikar** is a Solutions Architect at a large corporation, where he has been a key member of the Data Analytics and Solutions team since July 2016. He holds a Bachelor's degree in Data Science and Programming from the prestigious Indian Institute of Technology (IIT) Madras. Mr. Madkaikar's expertise lies at the intersection of applied mathematics and programming, with a strong focus on computational statistics, predictive analytics, deep learning, and cognitive science.

Mr. Madkaikar has contributed to the advancement of his field through published research. He is a co-author of the paper "3D Object Classification Using Bounding Box," presented at the 2018 Fourth International Conference on Computing Communication Control and Automation (ICCUBEA). His work explores the application of machine learning for analyzing complex 3D models in CAD drawings, with implications for industries such as automotive, robotics, and augmented reality. He has also co-authored research on intelligent alarm management systems for plant automation.